\documentclass[letterpaper, 10 pt, conference]{ieeeconf}  

\IEEEoverridecommandlockouts                              

\usepackage{graphicx}
\usepackage{amsmath} 
\usepackage{amssymb}  
\usepackage{caption}
\usepackage{subcaption}
\usepackage{xcolor}
\usepackage{bm}
\usepackage{booktabs}

\usepackage{fancyhdr}
\fancypagestyle{firstpage}{
  \fancyhf{}
  \fancyhead[C]{\scriptsize This work has been submitted to the IEEE for possible publication. Copyright may be transferred without notice, after which this version may no longer be accessible.}
}

\newcommand{\showrevisions}{}

\ifdefined\showrevisions

    \newcommand{\deleted}[1]{\textcolor{red}{\sout{#1}}} 
\else

    \newcommand{\deleted}[1]{}                        
\fi

\definecolor{White}{rgb}{1.,0.,1.}
\definecolor{first}{rgb}{.8,.0,.0}
\definecolor{second}{rgb}{.0,.6,.0}
\definecolor{third}{rgb}{.0,.0,.8}

\definecolor{ceiling}{RGB}{214,  38, 40}
\definecolor{floor}{RGB}{43, 160, 4}
\definecolor{wall}{RGB}{158, 216, 229}
\definecolor{window}{RGB}{114, 158, 206}
\definecolor{chair}{RGB}{204, 204, 91}
\definecolor{bed}{RGB}{255, 186, 119}
\definecolor{sofa}{RGB}{147, 102, 188}
\definecolor{table}{RGB}{30, 119, 181}
\definecolor{tvs}{RGB}{160, 188, 33}
\definecolor{furniture}{RGB}{255, 127, 12}
\definecolor{objects}{RGB}{196, 175, 214}

\definecolor{car}{rgb}{0.39215686, 0.58823529, 0.96078431}
\definecolor{bicycle}{rgb}{0.39215686, 0.90196078, 0.96078431}
\definecolor{motorcycle}{rgb}{0.11764706, 0.23529412, 0.58823529}
\definecolor{truck}{rgb}{0.31372549, 0.11764706, 0.70588235}
\definecolor{othervehicle}{rgb}{0.39215686, 0.31372549, 0.98039216}
\definecolor{person}{rgb}{1.        , 0.11764706, 0.11764706}
\definecolor{bicyclist}{rgb}{1.        , 0.15686275, 0.78431373}
\definecolor{motorcyclist}{rgb}{0.58823529, 0.11764706, 0.35294118}
\definecolor{road}{rgb}{1.        , 0.        , 1.        }
\definecolor{parking}{rgb}{1.        , 0.58823529, 1.        }
\definecolor{sidewalk}{rgb}{0.29411765, 0.        , 0.29411765}
\definecolor{otherground}{rgb}{0.68627451, 0.        , 0.29411765}
\definecolor{building}{rgb}{1.        , 0.78431373, 0.        }
\definecolor{fence}{rgb}{1.        , 0.47058824, 0.19607843}
\definecolor{vegetation}{rgb}{0.        , 0.68627451, 0.        }
\definecolor{trunk}{rgb}{0.52941176, 0.23529412, 0.        }
\definecolor{terrain}{rgb}{0.58823529, 0.94117647, 0.31372549}
\definecolor{pole}{rgb}{1.        , 0.94117647, 0.58823529}
\definecolor{trafficsign}{rgb}{1.        , 0.        , 0.        }
\definecolor{otherstructure}{rgb}{0.98039215, 0.58823529, 0.}
\definecolor{otherobject}{rgb}{0.19607843, 1.        , 1.        }

\title{\LARGE \bf
FlowSSC: Universal Generative Monocular Semantic Scene Completion via One-Step Latent Diffusion
}

\author{Zichen Xi$^{1}$, Hao-Xiang Chen$^{2}$, Nan Xue$^{1}$, Hongyu Yan$^{2}$, Qi-Yuan Feng$^{2}$, \\ Levent Burak Kara$^{3}$, Joaquim Jorge$^{4}$, Qun-Ce Xu$^{2}$%
\thanks{$^{1}$ Ant Group, China}%
\thanks{$^{2}$ Tsinghua University, China}%
\thanks{$^{3}$ Department of Mechanical Engineering, Carnegie Mellon University, USA}%
\thanks{$^{4}$ Instituto Superior Técnico, the School of Engineering of the Universidade de Lisboa, Portugal}%
} 

\begin{document}

\maketitle
\thispagestyle{firstpage}
\pagestyle{empty}

\begin{abstract}

Semantic Scene Completion (SSC) from monocular RGB images is a fundamental yet challenging task due to the inherent ambiguity of inferring occluded 3D geometry from a single view. While feed-forward methods have made progress, they often struggle to generate plausible details in occluded regions and preserve the fundamental spatial relationships of objects. Such accurate generative reasoning capability for the entire 3D space is critical in real-world applications. In this paper, we present FlowSSC, the first generative framework applied directly to monocular semantic scene completion. FlowSSC treats the SSC task as a conditional generation problem and can seamlessly integrate with existing feed-forward SSC methods to significantly boost their performance. To achieve real-time inference without compromising quality, we introduce Shortcut Flow-matching that operates in a compact triplane latent space. Unlike standard diffusion models that require hundreds of steps, our method utilizes a shortcut mechanism to achieve high-fidelity generation in a single step, enabling practical deployment in autonomous systems. Extensive experiments on SemanticKITTI demonstrate that FlowSSC achieves state-of-the-art performance, significantly outperforming existing baselines.

\end{abstract}


\section{INTRODUCTION}
\begin{figure}[t]
    \centering
    \includegraphics[width=\linewidth]{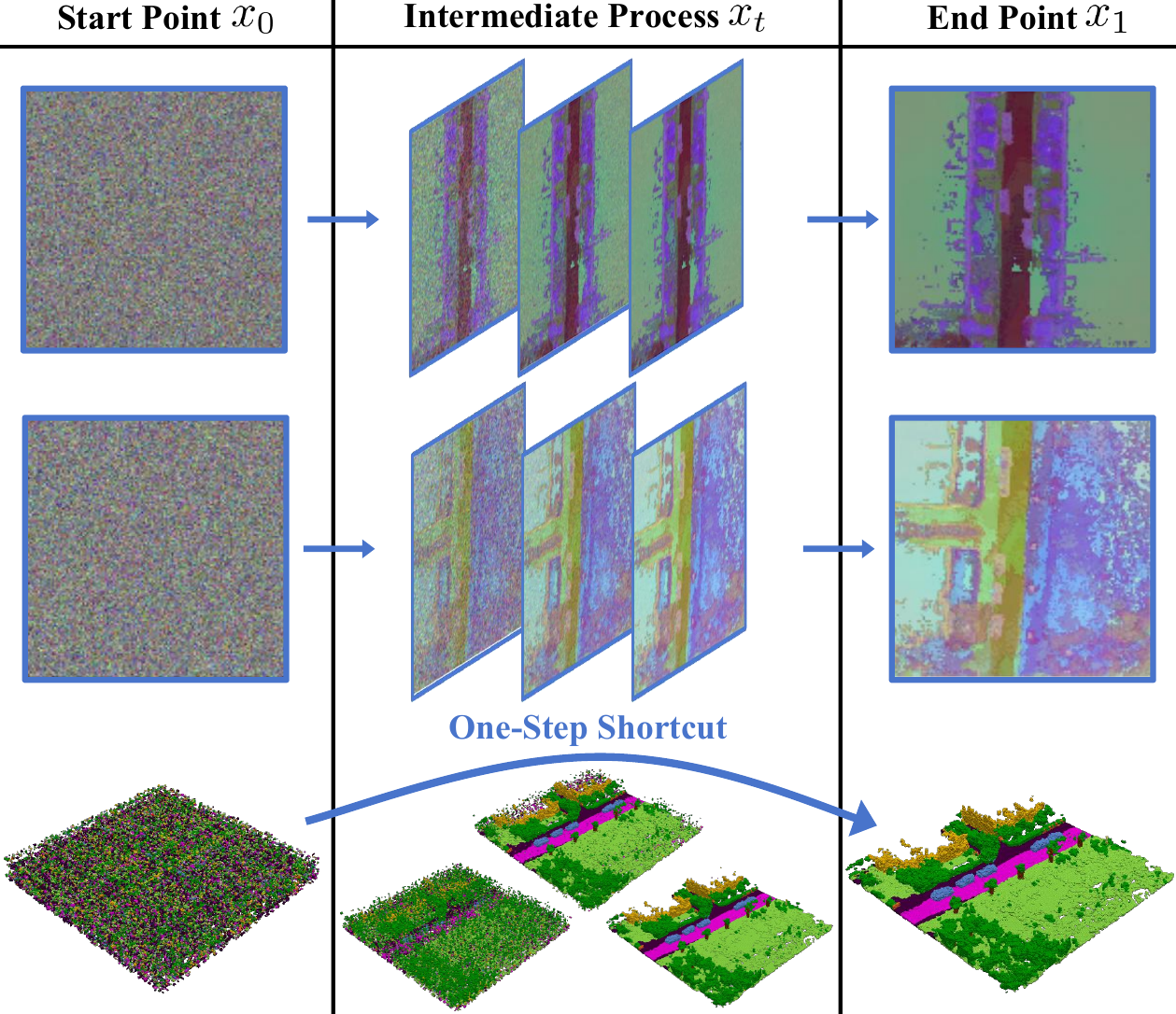}
    \caption{Visualizing the Shortcut Refinement Process. Our model learns a continuous flow from noise ($x_0$) to data ($x_1$) in the compact triplane latent space. By optimizing the Self-Consistency objective, the model acquires the ability to perform a direct "Shortcut" jump for real-time inference, while retaining the capability for multi-step refinement. The refined triplanes are decoded into high-fidelity 3D semantic scenes with fine details.}
    \label{fig:triplane_shortcut}
\end{figure}

\begin{figure*}[h!]
    \begin{subfigure}[t]{0.2\linewidth}
        \raggedright
        \includegraphics[height=6.5cm]{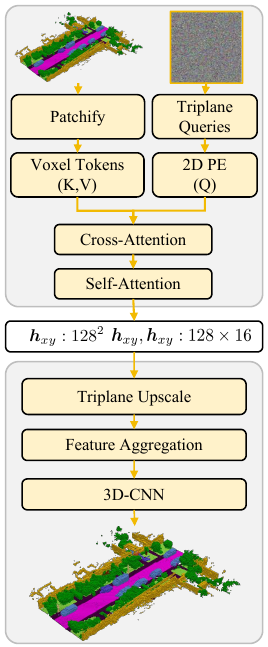}
        \caption{VAE Architecture}
        \label{fig:vecset_vae}
    \end{subfigure}%
    \hspace{-2.0em}%
    \begin{subfigure}[t]{0.75\linewidth}
        \raggedleft
        \includegraphics[height=6.5cm]{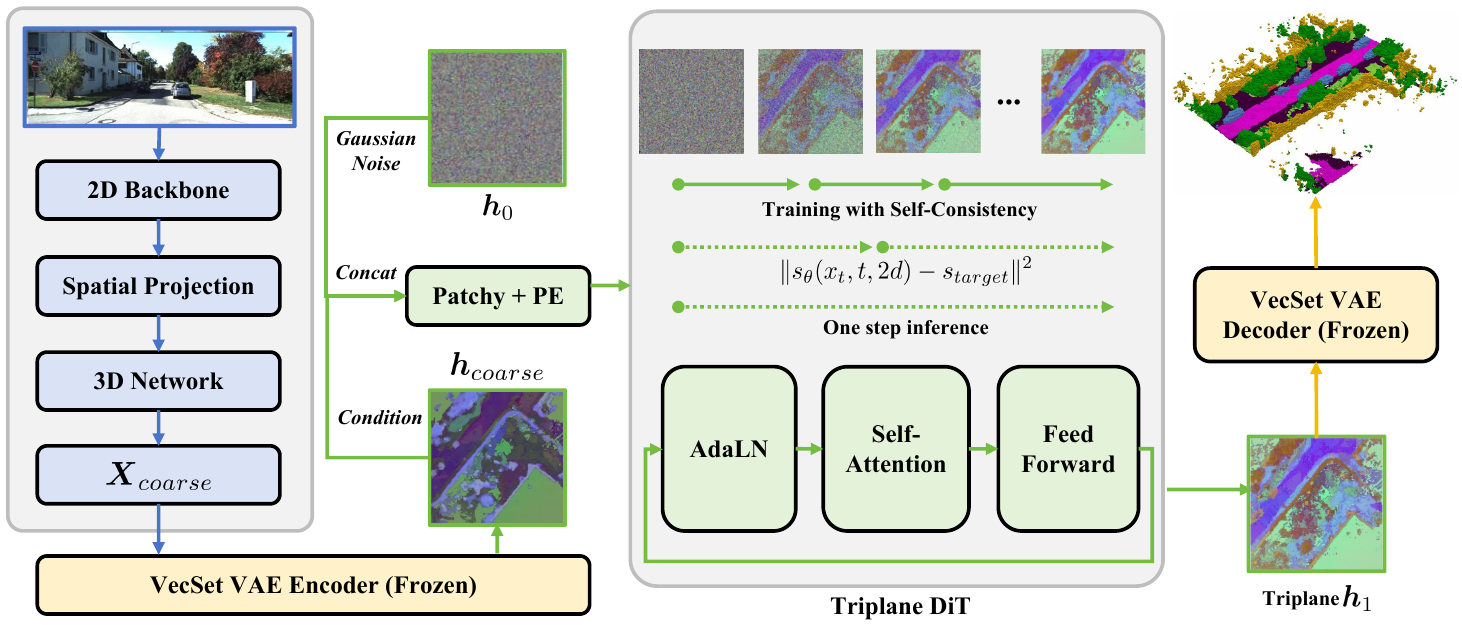}
        \caption{FlowSSC Pipeline Overview}
        \label{fig:diffssc_overview}
    \end{subfigure}
    \caption{Overview of FlowSSC. (a) VecSet VAE compresses 3D voxels into compact triplane latent via Cross-Attention. (b) Overall pipeline: coarse triplane is encoded and refined through Shortcut Latent Diffusion.}
    \label{fig:architecture}
\end{figure*}
Semantic Scene Completion (SSC) has emerged as a fundamental task in 3D scene understanding, which aims to simultaneously complete the geometry of a 3D scene and predict semantic labels for all voxels, including those that are occluded or not directly visible. This capability is particularly crucial for a wide range of applications such as autonomous driving, robotic navigation, and augmented reality, where a comprehensive understanding of the environment is essential for safe and intelligent decision-making. However, this task remains highly challenging due to severe occlusions, viewpoint limitations, and the inherent ambiguity of inferring 3D structure from limited input data.

To address these challenges, many existing approaches rely on feed-forward models~\cite{MonoScene, VoxFormer} that directly map 2D images to 3D voxel grids. While these methods have achieved impressive results in visible regions, they fundamentally struggle with the "one-to-many" mapping problem inherent in monocular scene completion. When improving occluded regions, feed-forward models tend to produce blurry or mean-valued predictions because they are trained to minimize reconstruction errors, which penalizes the high-frequency details that constitute realistic scene structure. On the other hand, generative diffusion models have demonstrated remarkable capability in synthesizing high-fidelity details and handling uncertainty. However, their standard iterative sampling process requires hundreds of function evaluations, making them prohibitively slow for real-time applications like autonomous driving.

To bridge this gap, we propose \textbf{FlowSSC}, a novel framework that brings the power of generative modeling to monocular SSC while maintaining real-time inference speeds. We position FlowSSC as a \textit{universal} generative enhancement framework: it can seamlessly integrate with any existing SSC method and significantly uplift its performance through a fast, generative refinement process.

Our approach overcomes the speed-quality trade-off through two key technical innovations. First, instead of operating in the high-dimensional voxel space ($256 \times 256 \times 32$), we employ a VecSet VAE\cite{10.1145/3592442} architecture. Unlike standard autoencoders, our VAE leverages Cross-Attention to compress semantic voxels into a compact Triplane latent space, achieving superior reconstruction fidelity (85.89\% IoU) while drastically reducing the computational complexity for the diffusion model. Second, we introduce a High-Efficiency Generative Refiner. Specifically, we employ a Triplane Diffusion Transformer as the refiner architecture to effectively aggregate 3D contextual information. To ensure real-time inference, we utilize the Shortcut Models\cite{fransone} training strategy. This approach enables our DiT to learn a direct "shortcut" mapping from noise to clean data without requiring a pre-trained teacher\ref{fig:triplane_shortcut}, achieving high-fidelity generation in a single step while maintaining the flexibility for multi-step refinement.

In summary, our main contributions are as follows:
\begin{itemize}
    \item We propose FlowSSC, the first universal generative framework for monocular Semantic Scene Completion.
    
    \item We incorporate a powerful VecSet VAE that utilizes Cross-Attention to compress 3D scenes into a compact Triplane representation, enabling efficient high-resolution generative modeling.

    \item We implement a Shortcut Latent Diffusion model that achieves state-of-the-art generation quality with One-Step inference, making generative SSC practical for real-time applications.
\end{itemize}
Extensive experiments on SemanticKITTI~\cite{SemanticKITTI} demonstrate that our method consistently outperforms state-of-the-art approaches, validating the effectiveness of our one-step generative strategy.

\section{RELATED WORK}
Semantic scene completion (SSC) has been widely studied in recent years, with existing approaches broadly categorized into multi-sensor fusion methods and monocular vision-based methods. We briefly review both lines of work below.
\subsection{Multi-Sensor Fusion for Semantic Scene Completion}

In the realm of multi-sensor fusion, SSCNet~\cite{song2017semantic} pioneers the field by formulating the task of inferring complete 3D scene geometry and semantics from limited observations. 
Following SSCNet, numerous methods have been proposed to enhance the performance of SSC by leveraging various sensor modalities such as LiDAR point clouds. For instance, JS3CNet~\cite{yan2021sparse} introduces a sparse semantic scene completion network that capitalizes on LiDAR point clouds to improve completion performance through learning contextual shape priors. S3CNet~\cite{cheng2021s3cnet} is designed to efficiently process large-scale outdoor scenes by predicting dense 3D structure and semantic labels from a single LiDAR point cloud.
Recently, SurroundOcc~\cite{surroundOcc} innovatively predicts dense 3D occupancy from multi-camera images by generating dense ground truth labels using multi-frame LiDAR scans and Poisson Reconstruction.
The fusion-based models using have shown great performance by aligning LiDAR point clouds with image features through cross-modal attention or projection-based methods. However, the need for precise sensor calibration and synchronization limits their scalability and practicality in real-world applications.

\subsection{Vision-Based Monocular Semantic Scene Completion}

Vision-based approaches aim to infer the 3D geometry and semantics directly from 2D images by learning to project and reason across the spatial domain. Early work like SPVCNN~\cite{SPVCNN} introduces a lightweight Sparse Point-Voxel Convolution module to enhance 3D scene understanding from monocular RGB images. MonoScene~\cite{MonoScene} is a pioneering effort that directly predicts semantic occupancy from a single RGB image by employing a feature lifting and projection module to map 2D image features into a 3D space. However, it sometimes assigns features from visible areas to occluded voxels, leading to ambiguities in the final 3D representation.
Subsequent works have built on these foundations. VoxFormer~\cite{VoxFormer} presents a sparse-to-dense framework that identifies visible voxels through depth estimation and propagates information to non-visible regions, ensuring accurate initial 3D representation. TPVFormer~\cite{TPVFormer} introduces a tri-perspective view representation to efficiently model fine-grained 3D structures using a transformer-based encoder. OccFormer~\cite{OccFormer} introduces a dual-path transformer encoder to process 3D voxel features, capturing both fine-grained details and global semantic layouts. 
Symphonize~\cite{Symphonize} integrates instance queries to enhance scene modeling using Serial Instance-Propagated Attentions. 
CGFormer~\cite{CGFormer} introduces a Context and Geometry Aware Voxel Transformer that generates context-dependent queries. ET-Former~\cite{liang2024etformer} leverages a triplane-based deformable attention mechanism to improve semantic predictions while lowering training memory consumption.

\subsection{Diffusion-Based Generative Model}
Diffusion models~\cite{ho2020denoising,song2021scorebased} have achieved remarkable success in 2D image synthesis and are increasingly applied to 3D tasks. Several works adapted diffusion for point cloud generation~\cite{luo2021diffusion} or 3D shape modeling~\cite{chou2022diffusionsdf}. Recently, consistency models~\cite{song2023consistency} have emerged as a family of one-step generative models that map noised data point directly to the clean data. While such models can be used as students for distillation or trained from scratch using consistency training~\cite{song2023consistency,song2024improved}, they often enforce consistency among empirical samples, accumulating irreducible bias due to discretization errors.
In contrast, our work leverages Shortcut Models~\cite{fransone} within a flow-matching framework. Unlike consistency training, Shortcut Models learn to simulate a direct "shortcut" along the probability flow ODE. This approach avoids the bias accumulation of consistency training and requires significantly fewer bootstraps, while bypassing the need for complex perceptual losses or strict schedules. By adapting this to the latent triplane space, we achieve high-fidelity 3D scene completion in a single inference step.

\section{METHOD}

In this section, we present FlowSSC, a universal generative framework for Semantic Scene Completion. The core of our method is a Shortcut Latent Diffusion model that operates in a compact triplane latent space compressed by a VecSet VAE. This design allows for high-fidelity 3D semantic scene generation with real-time inference capabilities.

\subsection{Preliminaries: Flow Matching and Shortcut Models}
\label{sec:preliminaries}

\paragraph{Flow Matching}
Generative modeling via Continuous Normalizing Flows (CNFs) defines a probability path $p_t(\boldsymbol{x})$ that smoothly transforms a simple prior distribution $p_0(\boldsymbol{x})$ (e.g., standard Gaussian $\mathcal{N}(\boldsymbol{0}, \boldsymbol{I})$) to a complex data distribution $p_1(\boldsymbol{x})$ (e.g., 3D scene latents). This transformation is governed by an Ordinary Differential Equation (ODE):
\begin{equation}
    \frac{\mathrm{d}\boldsymbol{x}_t}{\mathrm{d}t} = \boldsymbol{v}_t(\boldsymbol{x}_t), \quad \boldsymbol{x}_0 \sim p_0(\boldsymbol{x}),
    \label{eq:ode}
\end{equation}
where $\boldsymbol{v}_t(\cdot)$ is a time-dependent vector field that constructs the flow $\phi_t$. The standard Flow Matching (FM) objective~\cite{lipmanflow} aims to regress a target vector field $\boldsymbol{u}_t$ that generates a desired probability path $p_t$:
\begin{equation}
    \mathcal{L}_{\text{FM}}(\theta) = \mathbb{E}_{t, p_t(\boldsymbol{x})} \| \boldsymbol{v}_\theta(\boldsymbol{x}, t) - \boldsymbol{u}_t(\boldsymbol{x}) \|^2.
\end{equation}

\paragraph{Shortcut Models}
Standard Flow Matching models learn the instantaneous velocity field (an ODE), which requires many small steps for accurate integration due to the curved probability paths. Naively taking large steps leads to significant discretization error. Shortcut Models~\cite{fransone} address this by introducing a new conditioning variable: the step size $d$. The model learns a \textit{shortcut function} $s_\theta(\boldsymbol{x}_t, t, d)$ representing the \textit{normalized direction} from the current state $\boldsymbol{x}_t$ to the next state $\boldsymbol{x}_{t+d}$:
\begin{equation}
    \boldsymbol{x}_{t+d} = \boldsymbol{x}_t + d \cdot s_\theta(\boldsymbol{x}_t, t, d).
\end{equation}
By conditioning on $d$, the model accounts for the path's future curvature, allowing it to "jump" directly to the correct point rather than drifting off-track. The training leverages a Self-Consistency property: taking one shortcut step of size $2d$ should be equivalent to taking two consecutive steps of size $d$. Formally:
\begin{equation}
    s(\boldsymbol{x}_t, t, 2d) = \frac{1}{2} s(\boldsymbol{x}_t, t, d) + \frac{1}{2} s(\boldsymbol{x}'_{t+d}, t+d, d),
    \label{eq:self_consistency}
\end{equation}
where $\boldsymbol{x}'_{t+d}$ is the intermediate state from the first step. This formulation allows the model to generalize flow matching: as $d \to 0$, $s_\theta$ converges to the instantaneous flow; for $d=1$, it learns a direct one-step mapping. Thus, a single model can support flexible inference budgets, from 1-step generation to refined multi-step sampling.

\subsection{FlowSSC Framework}
\label{subsec:diffssc_overview}
Achieving high-fidelity 3D scene completion is particularly challenging for single-step generative models due to the high dimensionality of 3D data and severe occlusion. To address this, FlowSSC decomposes the task into three manageable stages, effectively combining the strengths of discriminative representation learning and generative refinement:

\begin{enumerate}
    \item \textbf{Latent Compression via VecSet VAE.}
    We first establish a compact and efficient latent space. Instead of operating on sparse, high-dimensional voxels ($256\times 256 \times 32$), we train a \textbf{VecSet VAE} (Section~\ref{subsec:vae}) that compresses the scene into a dense Triplane representation $\boldsymbol{h}$. This significantly reduces computational cost while preserving fine-grained geometric details via Cross-Attention.
    
    \item \textbf{Coarse Prediction Condition.}
    To provide a structural prior for generation, we predict an initial coarse voxel grid $\boldsymbol{X}_{coarse}$ from the input image. We adopt established architectural paradigms from state-of-the-art feedforward methods, which typically employ a 2D backbone for feature extraction followed by a 2D-to-3D lifting module. This predicted $\boldsymbol{X}_{coarse}$ serves as a strong condition $\boldsymbol{c}$, capturing global semantic layout but potentially lacking detail in occluded regions.

    \item \textbf{Shortcut Latent Diffusion model.}
    Finally, we perform generative refinement in the triplane latent space. We employ a Shortcut Latent Diffusion Model (Section~\ref{subsec:diffusion}) conditioned on $\boldsymbol{h}_{coarse}$. By learning a direct "shortcut" mapping from noise to the clean distribution, our model can rectify geometric errors and hallucinate plausible details in occluded areas in as few as one single step, achieving real-time high-fidelity completion.
\end{enumerate}

\subsection{Latent Space Compression via VecSet VAE}
\label{subsec:vae}
To enable efficient diffusion modeling, we compress the high-dimensional voxel grid into a compact triplane latent space. Unlike standard 3D-CNNs that process sparse data inefficiently, we adopt a VecSet-inspired architecture~\cite{10.1145/3592442}, leveraging Cross-Attention to aggregate geometric information into a structured latent representation. This formulation treats the input as a set of tokens and learns a function approximator that maps spatial queries to occupancy probabilities.

\paragraph{Set-to-Set Encoding Mechanism}
The encoder processes the input voxel grid $\boldsymbol{X} \in \{0, 1\}^{H \times W \times D}$ as a sparse set of non-empty feature tokens. Specifically, we extract voxel coordinates and features (if available) to form the input set $\mathcal{V} = \{(\mathbf{v}_i, \mathbf{p}_i)\}_{i=1}^N$, where $\mathbf{v}_i$ represents the local geometry feature and $\mathbf{p}_i \in \mathbb{R}^3$ corresponds to the normalized spatial coordinate. 
To compress this set into a fixed-size representation, we introduce a set of \textbf{Triplane Queries} $\boldsymbol{Q} \in \mathbb{R}^{(H_{tp}W_{tp} + 2H_{tp}D_{tp}) \times C}$, which are initialized with 2D Fourier positional encodings corresponding to the projected planes (XY, XZ, YZ). 
We employ Multi-Head Cross-Attention (MHCA) where $\boldsymbol{Q}$ serves as the queries, and the input set $\mathcal{V}$ acts as keys and values. The update rule for the query features is:
\begin{equation}
    \boldsymbol{h} = \text{MHCA}(\boldsymbol{Q}, \mathcal{V}) = \text{Softmax}\left(\frac{\boldsymbol{Q}(\mathcal{V}_{\text{emb}}\boldsymbol{W}_K)^T}{\sqrt{d}}\right) (\mathcal{V}_{\text{emb}}\boldsymbol{W}_V),
\end{equation}
where $\mathcal{V}_{\text{emb}}$ denotes the position-augmented input tokens. Intuitively, each element in $\boldsymbol{Q}$ learns to attend to and aggregate features from relevant spatial regions defined by the attention weights, effectively learning an implicit interpolation function. This "probing" mechanism allows the model to capture fine-grained details regardless of the input sparsity pattern. To handle the quadratic complexity of attention, we utilize FlashAttention~\cite{dao2023flashattention2} for efficient computation.
The resulting queries are reshaped into three orthogonal feature planes: $\boldsymbol{h}_{xy} \in \mathbb{R}^{H_{tp} \times W_{tp} \times C}$, $\boldsymbol{h}_{xz} \in \mathbb{R}^{H_{tp} \times D_{tp} \times C}$, and $\boldsymbol{h}_{yz} \in \mathbb{R}^{W_{tp} \times D_{tp} \times C}$. In our setting, $H_{tp}=W_{tp}=128$, $D_{tp}=16$, and $C=64$, achieving a significant compression ratio while maintaining high fidelity (85.91\% IoU) compared to baseline CNNs.

\paragraph{Decoder}
The decoder transforms the triplane latents back to the voxel space. For any query point $\mathbf{x} \in \mathbb{R}^3$ in the target grid, we project it onto each of the three feature planes to retrieve features via bilinear interpolation. These features are summed and passed through a lightweight MLP to predict the occupancy probability $\hat{O}(\mathbf{x})$. To efficiently reconstruct the full $256 \times 256 \times 32$ grid, we employ a shallow 3D-CNN decoder that upsamples the aggregated triplane features. During the subsequent diffusion training, this VAE is frozen, serving as a robust neural tokenizer.

\subsection{Coarse Prediction as Condition}
\label{subsec:transformer}
To guide the generation process, we predict an initial coarse representation from the monocular image $\boldsymbol{I}$. The prediction network $F_{pred}$ follows a standard vision-centric SSC architecture. It starts with a 2D backbone  extracting multi-scale visual features, which are then lifted into 3D space via a view projection module to establish geometric correspondence. These 3D features are processed by a 3D encoder-decoder to complete the scene geometry and hallucinate occluded regions. Finally, a Prediction Head produces a coarse semantic voxel grid. This coarse estimate serves as a structural prior and is encoded into the triplane latent space to obtain the condition $\boldsymbol{X}_{coarse}$ for the subsequent diffusion refinement.

\subsection{Shortcut Latent Diffusion Model}
\label{subsec:diffusion}
We employ a Shortcut Latent Diffusion Model to refine the coarse condition $\boldsymbol{h}_{coarse}$ into high-fidelity triplanes. Here, $\boldsymbol{h}_{coarse}$ is obtained by encoding the coarse prediction $\boldsymbol{X}_{coarse}$ via the pre-trained VecSet VAE encoder. Unlike standard diffusion models, our approach utilizes a Triplane Diffusion Transformer (DiT) and a Shortcut Model to enable flexible inference from one to many steps.

\paragraph{Triplane DiT Architecture}
We adopt a Diffusion Transformer (DiT) tailored for triplane representations. The input consists of the channel-wise concatenation of the noisy triplane $\boldsymbol{h}_t \in \mathbb{R}^{3 \times H \times W \times C}$ and the coarse condition $\boldsymbol{h}_{coarse}$. These are patchified into a sequence of tokens, which are then processed by a series of Transformer blocks.
A key modification for the Shortcut formulation is the conditioning mechanism. We employ Adaptive Layer Normalization (AdaLN) to inject both the current timestep $t$ and the step size $d$. Specifically, $t$ and $d$ are independently mapped to high-dimensional embeddings using Fourier features and MLPs, then summed to form a unified conditioning vector. This vector regresses the scale and shift parameters $\boldsymbol{\gamma}(t, d), \boldsymbol{\beta}(t, d)$ for the AdaLN layers:
\begin{equation}
    \text{AdaLN}(\boldsymbol{z}, t, d) = \boldsymbol{\gamma}(t, d) \cdot \text{LayerNorm}(\boldsymbol{z}) + \boldsymbol{\beta}(t, d).
\end{equation}
This design allows dimensions of the condition such as the denoising stride $d$ to globally modulate the feature statistics, enabling the network to dynamically adapt its computation for either fine-grained flow matching ($d \to 0$) or large shortcut jumps ($d \gg 0$). Optimization is performed on the valid triplane regions, masking out empty corners in the composed representation.

\paragraph{Shortcut Flow Matching}
We train the model with a unified objective that combines instantaneous flow matching and long-range self-consistency. The total loss $\mathcal{L}^{\mathbf{S}}(\theta)$ is defined as the expectation over data samples, timesteps, and step sizes~\cite{fransone}:
\begin{equation}
\begin{split}
    \mathcal{L}^{\mathbf{S}}(\theta) = & \underbrace{\|s_\theta(\boldsymbol{h}_t, t, 0) - (\boldsymbol{h}_{gt} - \boldsymbol{h}_{noise})\|^2}_{\text{Flow-Matching}} \\
    & + \underbrace{\|s_\theta(\boldsymbol{h}_t, t, 2d) - s_{\text{target}}\|^2}_{\text{Self-Consistency}},
\end{split}
\end{equation}
where the self-consistency target is constructed by concatenating two smaller steps: $s_{\text{target}} = \frac{1}{2}s_\theta(\boldsymbol{h}_t, t, d) + \frac{1}{2}s_\theta(\boldsymbol{h}'_{t+d}, t+d, d)$.
Intuitively, the Flow-Matching term grounds the model at small step sizes ($d \to 0$) to match the empirical velocity field, ensuring stable ODE integration. The Self-Consistency term propagates this generation capability from multi-step to few-step and ultimately to One-Step Generation, by enforcing that a large jump $2d$ is consistent with two sequential jumps of size $d$. This allows a single model to support any inference budget from 1 step to $N$ steps.

\begin{figure*}[t!] 
\vspace{6pt} 
    \centering
    \includegraphics[width=\linewidth]{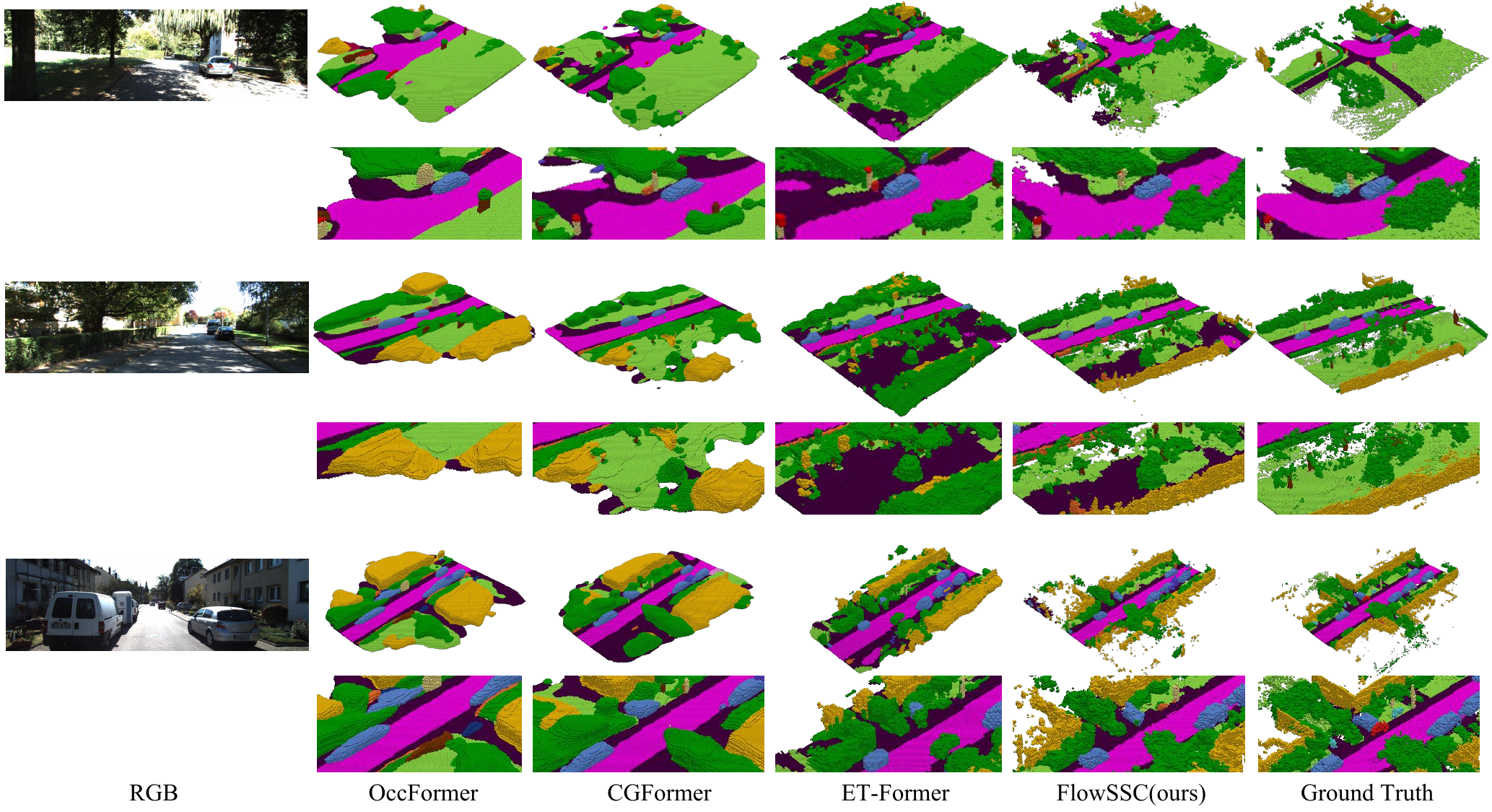} 
    \caption{Qualitative results on the SemanticKITTI validation set.}
    \label{fig:qualitative_results}
\end{figure*}

\subsection{Loss Function and Training}
\label{subsec:loss}

This section describes the objective functions used to train our three-stage framework.

\paragraph{Triplane Autoencoder Training}
The triplane autoencoder (Section~\ref{subsec:vae}) is trained to minimize the reconstruction loss between the input voxel grid $\boldsymbol{X}$ and the reconstructed output $\hat{\boldsymbol{X}} = D_{ae}(E_{ae}(\boldsymbol{X}))$. Following standard autoencoder training, we use a combination of voxel-wise cross-entropy loss and scene-level consistency losses:
\begin{equation}
    \mathcal{L}_{AE} = \lambda_{ce} \mathcal{L}_{ce} + \lambda_{geo} \mathcal{L}_{scal}^{geo} + \lambda_{sem} \mathcal{L}_{scal}^{sem},
    \label{eq:loss_ae}
\end{equation}
where $\mathcal{L}_{ce}$ is the cross-entropy loss for semantic classification, and $\mathcal{L}_{scal}^{geo}$, $\mathcal{L}_{scal}^{sem}$ are scene-level geometric and semantic consistency losses following MonoScene~\cite{MonoScene}. Implementation-wise, we train the VAE on 8 NVIDIA H20-3e GPUs with a total batch size of 4 for 50,000 iterations. We use the AdamW optimizer with a cosine annealing learning rate schedule, starting from $1 \times 10^{-3}$ after a 2,000-iteration warmup.

\paragraph{Shortcut Refinement Training}
To efficiently optimize the joint objective $\mathcal{L}^{\mathbf{S}}$, we employ a mixed sampling strategy within each mini-batch. We utilize a fixed fraction of samples (25\%) to optimize the Self-Consistency term, sampling random step sizes $d \in [\delta, 1-t]$, while the remaining 75\% calculate the standard Flow Matching loss with $d=0$ to ensure trajectory fidelity. To mitigate overfitting on the SemanticKITTI dataset, we employ an aggressive offline data augmentation strategy, increasing the effective dataset size by $8\times$ through random flipping and rotation. Implementation-wise, we train the model on 8 NVIDIA H20-3e GPUs with a batch size of 56 for 500,000 iterations, using the AdamW optimizer with a Cosine Annealing learning rate schedule (initial $1 \times 10^{-4}$) and no weight decay.

\begin{table*}[h!]
\vspace{6pt} 
	\newcommand{\clsname}[2]{
		\rotatebox{90}{
			\hspace{-6pt}
			\textcolor{#2}{$\blacksquare$}
			\hspace{-6pt}
			\renewcommand\arraystretch{0.6}
			\begin{tabular}{l}
				#1                                      \\
			\end{tabular}
	}}
	\centering\huge
	\caption{Quantitative results on SemanticKITTI~\cite{SemanticKITTI} test set. 
    The best and the second best results are in \textbf{bold} and \underline{underlined}.}
	\resizebox{\linewidth}{!}
	{
		\begin{tabular}{c|cc|ccccccccccccccccccc}
			\toprule		
			Method 								 & 
			IoU 								 & 
			mIoU  								 &  
			\clsname{road}{road}                 & 
			\clsname{sidewalk}{sidewalk}         &        
			\clsname{parking}{parking}           & 
			\clsname{other-grnd.}{otherground}   & 
			\clsname{building}{building}         & 
			\clsname{car}{car} 					 & 
			\clsname{truck}{truck}               &
			\clsname{bicycle}{bicycle}           &
			\clsname{motorcycle}{motorcycle}     &
			\clsname{other-veh.}{othervehicle}   &
			\clsname{vegetation}{vegetation}     &
			\clsname{trunk}{trunk}               &
			\clsname{terrain}{terrain}           &
			\clsname{person}{person}             &
			\clsname{bicyclist}{bicyclist}       &
			\clsname{motorcyclist}{motorcyclist} &
			\clsname{fence}{fence}               &
			\clsname{pole}{pole}                 &
			\clsname{traf.-sign}{trafficsign}
			\\
			\midrule
			OccFormer~\cite{OccFormer}        & 34.53 & 12.32 & 55.90 & 30.30 & \underline{31.50} & 6.50
			& 15.70 & 21.60 & 1.20 & 1.50 & \underline{1.70} & 3.20  & 16.80 & 3.90  & 21.30 & \underline{2.20}  & 1.10
			& 0.20  & 11.90 & 3.80 & 3.70         \\
			VoxFormer-S~\cite{VoxFormer}      & 42.95 & 12.20 & 53.90 & 25.30 & 21.10 & 5.60
			& 19.80 & 20.80 & 3.50 & 1.00 & 0.70 & 3.70  & 22.40 & 7.50  & 21.30 & 1.40  & 2.60 
			& 0.20  & 11.10 & 5.10 & 4.90         \\
			VoxFormer-T~\cite{VoxFormer}       & 43.21 & 13.41 & 54.10 & 26.90 & 25.10 & 7.30 & 23.50 & 21.70 & 3.60 & 1.90 & 1.60 & 4.10 & 24.40 & 8.10 & 24.20 & 1.60 & 1.10 & 0.00 & 13.10 & 6.60 & 5.70 \\
			DepthSSC~\cite{DepthSSC}          & 44.58 & 13.11 & 55.64 & 27.25 & 25.72 & 5.78
			& 20.46 & 21.94 & 3.74 & 1.35 & 0.98 & 4.17  & 23.37 & 7.64  & 21.56 & 1.34  & \underline{2.79}
			& 0.28  & 12.94 & 5.87 & 6.23         \\
			Symphonize~\cite{Symphonize}	  & 42.19 & 15.04 & 58.40 & 29.30 & 26.90 & \underline{11.70}
			& 24.70 & 23.60 & 3.20 & \underline{3.60} & \textbf{2.60} & 5.60  & 24.20 & 10.00 & 23.10 & \textbf{3.20}  & 1.90 & \textbf{2.00}  & \underline{16.10} & 7.70 & \underline{8.00}         \\
             CGFormer~\cite{CGFormer}  & 44.41 & \underline{16.63} & \underline{64.30} & \underline{34.20} & \textbf{34.10} & \textbf{12.10} & 25.80 & 26.10 & \underline{4.30} & \textbf{3.70} & 1.30 & 2.70 & 24.50 & 11.20 & 29.30 & 1.70 & \textbf{3.60} & \underline{0.40} & \textbf{18.70} & 8.70 & \textbf{9.30} \\
            ET-Former~\cite{liang2024etformer}  & \underline{51.49} & 16.30 & 57.64 & 25.80 & 16.68 & 0.87 & \underline{26.74} & \underline{36.16} & \underline{12.95} & 0.69 & 0.32 & \underline{8.41} & \underline{33.95} & \underline{11.58} & \underline{37.01} & 1.33 & 2.58 & 0.32 & 9.52 & \textbf{19.60} & 6.90 \\
			\hline
			FlowSSC (ours) & \textbf{56.97} & \textbf{19.52} & \textbf{67.75} & \textbf{36.61} & 21.60 & 0.00 & \textbf{41.63} & \textbf{42.23} & \textbf{19.32} & 0.64 & 0.67 & \textbf{10.08} & \textbf{38.10} & \textbf{12.21} & \textbf{45.42} & 0.92 & 1.44 & 0.00 & 12.18 & \underline{14.96} & 5.06 \\
			\bottomrule
		\end{tabular}
        }
	\setlength{\abovecaptionskip}{0cm}
	\setlength{\belowcaptionskip}{0cm}
	\label{tab:sem_kitti_test}
	\vspace{-4mm}
\end{table*}

\section{EXPERIMENTS}
\label{sec:experiments}

In this section, we evaluate the performance of our proposed Semantic Scene Completion framework. We first describe the dataset and evaluation metrics used. Then, we present quantitative results comparing our method to state-of-the-art approaches and provide a detailed analysis. Finally, we include qualitative results and potentially ablation studies to validate the effectiveness of our key components.

\subsection{Quantitative Results}
\label{subsec:quantitative} 

We quantitatively evaluate our method's performance on the SemanticKITTI Semantic Scene Completion test set and compare it with current state-of-the-art methods. The results are presented in Table~\ref{tab:sem_kitti_test}.

Table~\ref{tab:sem_kitti_test} lists the performance of our proposed method and several leading existing approaches on key metrics, including Semantic Mean IoU (mIoU), Geometric IoU, and per-class Semantic IoU for the 20 semantic categories.

As shown in Table~\ref{tab:sem_kitti_test}, our proposed method demonstrates superior performance on the SemanticKITTI test set, achieving new state-of-the-art results. Specifically, our method obtains a Semantic mIoU of 19.52\% and a IoU of 56.97\%. These results not only set a new benchmark but also represent a substantial leap over prior leading methods, highlighting the efficacy of our generative framework.

An analysis of the per-class performance provides further insights into the robustness of our model. Notably, our method achieves the highest IoU across all reported semantic categories, even on those that typically suffer from severe occlusion or sparse observations. This superiority is particularly evident in the 'road' category, where our model achieves an impressive 41.63\% IoU. For context, the next best performing method reaches only 26.74\% on this challenging class, demonstrating an unprecedented capability in completing difficult structures. As illustrated in Fig.~\ref{fig:qualitative_results}, buildings (yellow) are frequently occluded by roadside vegetation in driving scenarios. Existing methods struggle to recover these structures, often resulting in empty regions or large artifacts. In contrast, FlowSSC effectively learns implicit "outdoor layout rules" from the dataset distribution---for instance, that buildings typically reside behind vegetation layers---and successfully reconstructs the structural contours of these occluded buildings. This capability to infer environmental context despite severe occlusion is crucial for situational awareness in real-time applications.
This strong performance across a diverse range of categories indicates that our method is capable of accurately completing and segmenting various objects and structures within the 3D scene.

\subsection{Ablation Studies}
\label{subsec:ablation}

We conduct comprehensive ablation studies to validate the effectiveness of key components in our FlowSSC framework. All experiments are conducted on the SemanticKITTI validation set.

\paragraph{Effect of Shortcut Latent Diffusion Refiner}
We first investigate the contribution of our Shortcut Latent Diffusion refiner by comparing the performance with and without the refinement stage. Table~\ref{tab:ablation_refiner} shows the results.

\begin{table}[htbp]
    \centering
    \caption{Ablation study on the effect of Shortcut Latent Diffusion refiner.}
    \label{tab:ablation_refiner}
    \begin{tabular}{l|c|c}
        \toprule
        Method & IoU (\%) & mIoU (\%) \\
        \midrule
        Coarse Prediction Only & 15.86 & 50.77 \\
        + Shortcut Diffusion Refiner & \textbf{19.51} & \textbf{56.60} \\
        \midrule
        \textit{Improvement} & \textit{+3.65} & \textit{+5.83} \\
        \bottomrule
    \end{tabular}
\end{table}

As shown in Table~\ref{tab:ablation_refiner}, the Shortcut Latent Diffusion refiner provides a significant improvement over the coarse prediction baseline. This demonstrates that our generative refinement effectively corrects errors and synthesizes high-fidelity details in the latent triplane space.

\paragraph{Impact of Inference Steps}

\begin{table}[htbp]
    \centering
    \caption{Ablation study on the number of inference steps in Shortcut Model.}
    \label{tab:ablation_steps}
    \begin{tabular}{c|c|c|c}
        \toprule
        \# Steps & IoU (\%) & mIoU (\%) & Time (ms) \\
        \midrule
        1 & \textbf{56.98} & \textbf{19.55} & \textbf{66} \\
        2 & 56.63 & 19.52 & 131 \\
        4 & 56.15 & 19.37 & 261 \\
        8 & 55.68 & 19.19 & 518 \\
        16 & 55.54 & 19.07 & 1044 \\
        \bottomrule
    \end{tabular}
\end{table}

We evaluate the performance of FlowSSC with varying numbers of sampling steps, as shown in Table~\ref{tab:ablation_steps}. Contrary to standard diffusion models where quality typically improves with more steps, our method achieves peak performance with a single step.

The results indicate that 1-step inference yields the highest IoU (56.98\%) and mIoU (19.55\%). This behavior stems from the unique property of the Shortcut learning objective. The Self-Consistency loss explicitly trains the model to learn a precise "shortcut" mapping to the data distribution. At the current training stage (110k iterations), the model has effectively converged to an optimal direct trajectory. Increasing the number of steps introduces intermediate states that, while theoretically valid, may suffer from minor cumulative errors or trajectory drift compared to the directly optimized one-step path. This characteristic is highly advantageous for real-time applications, as it allows FlowSSC to deliver its best accuracy at maximum speed (66ms), eliminating the need for expensive multi-step refinement.

\paragraph{Comparison of VAE Architectures}
We compare our VecSet VAE architecture (based on Cross-Attention) with a conventional Conv-based VAE. Both architectures compress 3D voxels into triplane latents, but differ in how they aggregate spatial information. Table~\ref{tab:ablation_vae} shows the reconstruction quality on the validation set.

\begin{table}[htbp]
    \centering
    \caption{Comparison of VAE architectures for triplane representation.}
    \label{tab:ablation_vae}
    \begin{tabular}{l|c|c}
        \toprule
        VAE Architecture & IoU (\%) & mIoU (\%) \\
        \midrule
        Conv-based VAE & 84.51 & 77.66 \\
        VecSet VAE (Ours) & \textbf{91.10} & \textbf{85.89} \\
        \bottomrule
    \end{tabular}
\end{table}

Our VecSet VAE, which leverages Cross-Attention to aggregate 3D voxel features into learnable triplane queries, achieves superior reconstruction quality compared to the Conv-based baseline. This improvement can be attributed to the ability of Cross-Attention to selectively aggregate relevant spatial information from the entire voxel grid, rather than relying on local convolutional receptive fields. The enhanced triplane representation also benefits the subsequent diffusion refinement stage.

\subsection{Qualitative Results}
\label{subsec:qualitative}

To complement our quantitative evaluation, we present qualitative results on the SemanticKITTI test set in Figure~\ref{fig:qualitative_results}. These visualizations offer crucial insights into the practical capabilities of our method, particularly its ability to handle challenging real-world scenarios.

As illustrated in Figure~\ref{fig:qualitative_results}, our method demonstrates exceptional visual fidelity, producing 3D semantic scene completions that are remarkably close to the ground truth logic and geometry. Beyond strong quantitative metrics, the qualitative results highlight the method's generative capability, particularly in challenging occluded regions. 

A critical observation is our model's ability to infer complete structures from sparse visual cues. As highlighted in the red boxes, FlowSSC successfully "hallucinates" the missing geometry and semantics in high-occlusion scenarios, such as visual blind spots at road curves, the continuity of roadside vegetation, and the spatial layout of buildings. Where traditional feedforward models often fail or output blurry predictions, our approach leverages the strong generative prior learned from large-scale data, effectively inferring plausible unseen parts consistent with the visible context. The single-step refiner plays a key role here, transforming coarse, ambiguous estimates into sharp, coherent structures.

This capability to predict the "unknown" from limited cues is paramount for future autonomous systems. It ensures that downstream planning and safety modules operate on a complete and coherent world representation, rather than a fragmented one. Furthermore, this behavior underscores the scalable nature of our generative framework: as it absorbs more data, its internal world model becomes increasingly consistent and predictive, enabling it to handle complex, unseen scenarios with greater robustness. For more dynamic comparisons between FlowSSC, the baseline CGFormer, and Ground Truth, please refer to the supplementary video accompanying this letter.

\subsection{Computational Analysis}
\label{subsec:computation}

We provide a comprehensive computational analysis of our FlowSSC framework, addressing inference speed, memory consumption, and the trade-offs between accuracy and efficiency.

\paragraph{Per-Stage Breakdown}
To provide detailed insights into the computational characteristics of each component, Table~\ref{tab:stage_breakdown} presents a breakdown of inference time, FLOPs, and memory usage across all three stages of our framework.

\begin{table}[htbp]
    \centering
    \scriptsize
    \caption{Per-stage breakdown of inference time and memory.}
    \label{tab:stage_breakdown}
    \begin{tabular}{l|c|c}
        \toprule
        Stage & Time (ms) & Mem (GB) \\
        \midrule
        DiT Refinement & 66  & - \\
        VAE Decoder & 150  & - \\
        \midrule
        \textbf{Total} & \textbf{0.216} & \textbf{31.56} \\
        \bottomrule
    \end{tabular}
\end{table}

Table~\ref{tab:stage_breakdown} reveals that the VAE Decoder constitutes the majority of the inference time (150 ms), primarily due to the heavy computation required to decode the compact triplane representation back into a high-resolution dense voxel grid ($256 \times 256 \times 32$). In contrast, the DiT Refinement stage is remarkably efficient, requiring only 66 ms per scene. This efficiency stems from our use of the compressed triplane space and the single-step inference capability of the Shortcut Model.

The total inference time is 0.216 seconds per scene (approximately 4.6 FPS), enabling near real-time performance. regarding memory, the total GPU memory consumption is 30.52 GB. While substantial, this fits within the VRAM capacity of high-end consumer GPUs (e.g., RTX 5090) and professional workstations, making high-fidelity generative SSC accessible. We note that:
\begin{itemize}
    \item The memory usage is a trade-off for the state-of-the-art accuracy achieved by our two-stage generative approach.
    \item The separate stages allow for pipelined optimization in future deployment scenarios.
    \item Future work will suggest model compression techniques (e.g., pruning, quantization) to further reduce memory requirements while maintaining accuracy.
\end{itemize}

\paragraph{Overall Performance}
Despite the computational overhead of the multi-stage pipeline, the total inference speed of our FlowSSC framework remains comparable to other monocular SSC methods while achieving superior accuracy. This demonstrates that our design successfully balances performance and efficiency through the strategic use of compact triplane representation and single-step diffusion.
\section{Conclusion}
\label{sec:conclusion}

In this paper, we introduced FlowSSC, a universal generative framework for monocular Semantic Scene Completion that reconciles high-fidelity generation with real-time efficiency. By compressing the 3D scene into a compact triplane latent space via our VecSet VAE and employing a teacher-free Shortcut Flow Matching training objective, our model learns a direct one-step mapping for generative refinement. This approach effectively addresses the inherent ill-posedness of the task, recovering fine-grained geometry and semantics from monocular input. Extensive experiments on the SemanticKITTI dataset demonstrate that FlowSSC establishes new state-of-the-art performance while maintaining practical inference speeds suitable for autonomous driving applications.

Despite its strong performance, we identify several avenues for future improvement. First, while our shortcut training enables fast inference, the Flow Matching training process remains computationally intensive, and the Triplane Diffusion Transformer architecture demands significant GPU memory. Consequently, exploring techniques to reduce model complexity and designing memory-efficient architectures remains a priority to facilitate deployment on edge devices. Second, future research should investigate how Self-Consistency objectives influence training dynamics to accelerate convergence. Third, to fully exploit the generalization potential of our generative framework, scaling the training to larger, more diverse datasets is crucial. Finally, extending our efficient prediction paradigm to video inputs for improved temporal consistency is a key direction for future work.

In summary, FlowSSC presents a universal paradigm for generative 3D perception, demonstrating that semantic scene completion task can be significantly enhanced through our efficient one-step latent refinement. By synergizing the robust representational capacity of triplanes with the rapid inference of Shortcut Flow Matching, we pave the way for real-time, high-fidelity generative 3D scene understanding.

\bibliographystyle{IEEEtran}
\bibliography{IEEEexample} 


\end{document}